\documentclass{article}
\usepackage{ijcai25}

\usepackage{times}
\usepackage{soul}
\usepackage{url}
\usepackage[hidelinks]{hyperref}
\usepackage[utf8]{inputenc}
\usepackage[small]{caption}
\usepackage{graphicx}
\usepackage{amsmath}
\usepackage{amsthm}
\usepackage{multirow}
\usepackage{amssymb}

\usepackage{tabularx}
\usepackage{array}
\usepackage{multirow}
\usepackage{booktabs}
\usepackage{algorithm}
\usepackage{algorithmic}
\usepackage[switch]{lineno}
\usepackage{threeparttable}
\title{Surface Vision Mamba: Leveraging Bidirectional State Space Model 
for Efficient Spherical Manifold Representation}

\author{
Rongzhao He\textsuperscript{\rm 1}\and
Weihao Zheng\textsuperscript{\rm 1}\thanks{Corresponding author.}\and
Leilei Zhao\textsuperscript{\rm 2}\and
Ying Wang\textsuperscript{\rm 1}\and
Dalin Zhu\textsuperscript{\rm 3}\and\\
Dan Wu\textsuperscript{\rm 4}\footnotemark[1]\and
Bin Hu\textsuperscript{\rm 1}\footnotemark[1]
\\
\affiliations
$^1$Lanzhou University\\
$^2$Harbin Institute of Technology\\
$^3$Gansu Maternity and Child-care Hospital\\
$^4$Zhejiang University\\
\emails
\{herongzhao23, zhengweihao, bh\}@lzu.edu.cn,
danwu.bme@zju.edu.cn,
}
\begin{document}

\maketitle

\begin{abstract}
    Attention-based methods have demonstrated exceptional performance
    in modelling long-range dependencies on spherical cortical surfaces,
    surpassing traditional Geometric Deep Learning (GDL) models.
    However, their extensive inference time and high memory demands pose
    challenges for application to large datasets with limited computing
    resources. Inspired by the state space model in computer vision,
    we introduce the attention-free Vision Mamba (Vim) to spherical surfaces,
    presenting a domain-agnostic architecture for analyzing data on spherical
    manifolds. Our method achieves surface patching by representing spherical
    data as a sequence of triangular patches derived from a subdivided
    icosphere. The proposed Surface Vision Mamba (SiM) is evaluated on
    multiple neurodevelopmental phenotype regression tasks using cortical
    surface metrics from neonatal brains. Experimental results demonstrate
    that SiM outperforms both attention- and GDL-based methods,
    delivering 4.8 times faster inference and achieving 91.7\% lower memory
    consumption compared to the Surface Vision Transformer (SiT) under the
    Ico-4 grid partitioning. Sensitivity analysis further underscores the 
    potential of SiM to identify subtle cognitive developmental patterns. 
    The code is available at \textit{\url{https://github.com/Rongzhao-He/surface-vision-mamba}}.
\end{abstract}

\section{Introduction}

% The {\it IJCAI--25 Proceedings} will be printed from electronic
% manuscripts submitted by the authors. These must be PDF ({\em Portable
%         Document Format}) files formatted for 8-1/2$''$ $\times$ 11$''$ paper.

Many methods have been developed for traditional 
Euclidean space data, such as Convolution Neural 
Networks (CNNs) and attention-based~\cite{dahan2022surface,zhao2024attention} approaches. 
CNNs use a regular convolutional kernel to slide 
over the input data, calculating the weighted sum 
at each location, while attention-based methods 
treat the data as a sequence of patches. However, 
few models exist for non-Euclidean data 
consist of graph, manifold and hyperbolic space 
data which have more complex geometries and 
distance metrics. These types of data are 
typically crucial in domains such as neuroscience, 
social network analysis, and theoretical physics, 
where their unique structures provide rich but 
underutilized information.

\begin{figure}[!t]
    \centering
    \includegraphics[width=1\columnwidth]{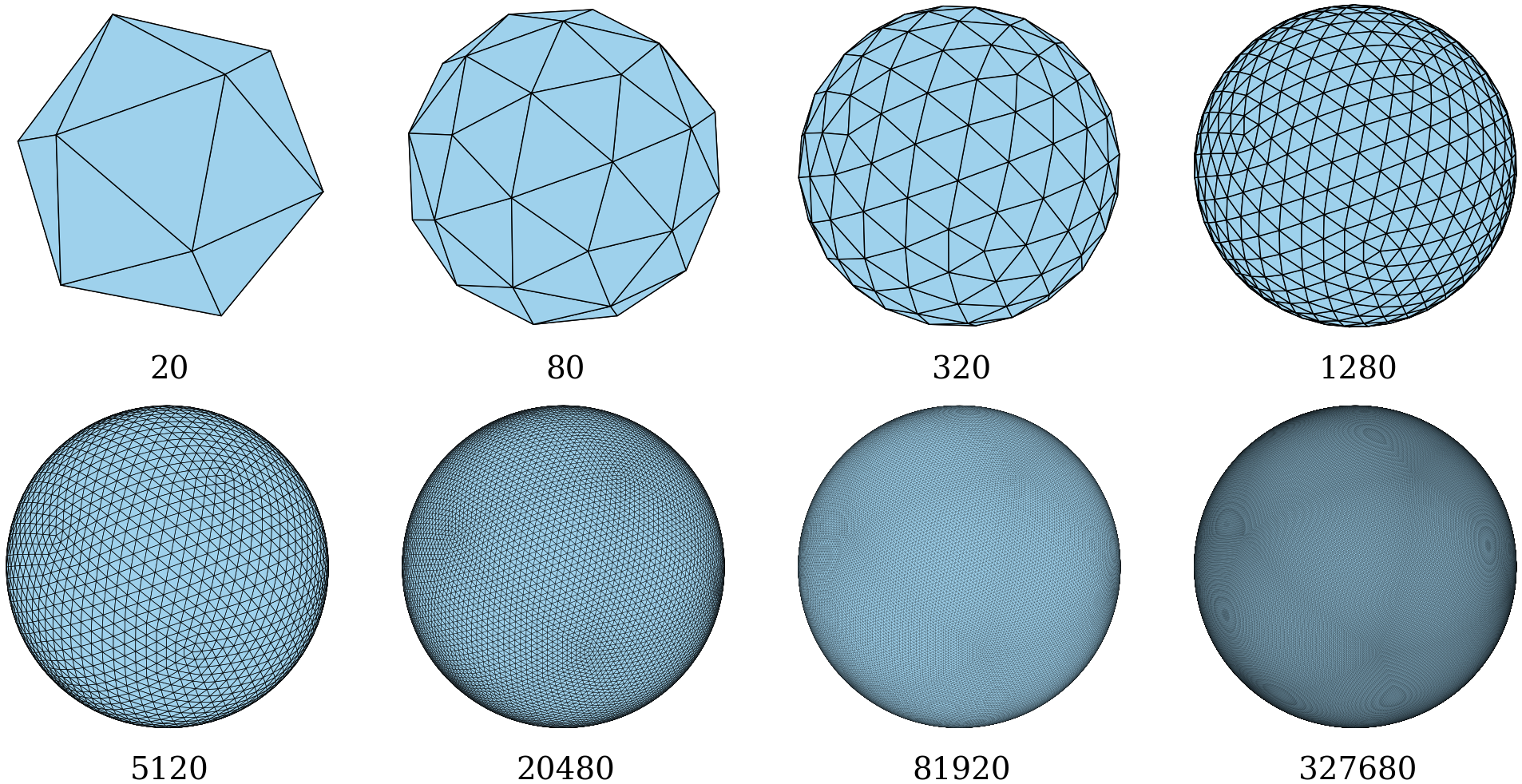}
    \caption{Representative icosahedron discretized spherical surfaces with sequential subdivisions. The number of faces of each spherical surface is denoted under the surface.}
    \label{fig1}
\end{figure}

Existing methods for processing non-Euclidean data can be 
broadly categorized into attention- and Geometric Deep 
Learning (GDL)-based~\cite{fawaz2021benchmarking,vosylius2020geometric} methods. Attention-based methods 
are effective in capturing long-range dependencies but are 
constrained in resource limited situations due to the 
quadratic complexity of the attention mechanism concerning 
sequence length, leading to higher memory consumption 
and slower inference time. Conversely, GDL-based methods, 
which operate directly on non-Euclidean data, are effective 
in handling complex geometric topology structure and 
distance metrics. However, they fail to extract global 
patterns, especially when applied to large-scale and 
highly intricate data, resulting in diminished performance. 
Thus, a key challenge for processing non-Euclidean data lies 
in improving efficiency while maintaining relatively 
excellent performance.

With the emergence of State Space Models (SSMs)~\cite{gu2021efficiently}, 
traditional sequence modeling methods have been revitalized, demonstrating 
promising capabilities for efficient representation learning. 
A recent variant Mamba~\cite{gu2023mamba}, has significantly surpassed 
traditional SSMs by integrating a selective scan mechanism that adapts 
parameters based on input and using a hardware-aware algorithm to 
parallelize scanning, thereby reducing memory I/O for more efficient inference. 
Motivated by ViT~\cite{dosovitskiy2020image}, ViG~\cite{han2022vision}, ~\cite{zhu2024vision} 
adapted Mamba to computer vision, introducing a bidirectional SSM structure to 
address direction-sensitive challenges, termed Vision Mamba (Vim).

Non-Euclidean data, particularly spherical cortical 
surface data, is characterized by high resolution, rich 
features, and intricate geometric shapes, as the cortical 
surface is inherently a high-dimensional manifold. While 
these data provide valuable insights into neurodevelopment, 
their effective representation poses a formidable challenge, 
often requiring a balance between performance and computational 
efficiency. Inspired by the efficiency of Vim, we extend 
its application to cerebral cortex analysis—an important 
yet underexplored area—by proposing Surface Vision Mamba (SiM). 
To adapt SiM to the unique characteristics of cortical 
surface data, we adjusted the input sequence length using 
various surface patching methods, as illustrated in Figure \ref{fig1}.

The main contributions of this study can be summarized as follows:
\begin{enumerate}
\item We introduce SiM, an adaptation of Vim, as a generic backbone network for analyzing data mapped onto genus-zero surfaces.

\item Leveraging the suitability of Mamba for tasks with \textit{long-sequence} and \textit{autoregressive} characteristics~\cite{yu2024mambaout}, we explore the impact of varying input sequence length on surface data in non-Euclidean space. We further implement autoregressive pretraining to validate the effectiveness of this approach.

\item Extensive experiments on three neurodevelopmental phenotype regression tasks, including the prediction of postmenstrual age (PMA) and long-term language and motor outcomes, demonstrate that our proposed SiM achieves promising performance compared to attention- and GDL-based models and is 4.8$\times$ faster than SiT and saves 91.7\% GPU memory when performing batch inference under the Ico-4 grid partitioning.

\end{enumerate}
% The main contributions of this study can be summarized as follows:
\section{Related Work}

% \LaTeX{} and Word style files that implement these instructions
% can be retrieved electronically. (See Section~\ref{stylefiles} for
% instructions on how to obtain these files.)

\subsection{Geometric Deep Learning}
Geometric Deep Learning (GDL) has emerged as a powerful 
tool for analyzing irregular geometries. While traditional 
CNNs specialize in processing Euclidean data, such as images, 
they are less effective for irregular 
data (e.g., cortical surfaces). GDL models extend CNNs to 
non-Euclidean domains, enabling the capture of intricate 
topological and geometric properties of the cortex. While 
these models excel in capturing local features, they often 
face challenges in learning long-range dependencies due to 
high computational costs or inherent architectural limitations, 
limiting their ability to model more complex relationships. 
Systematic comparisons of various GDL methods, such as 
MoNet~\cite{monti2017geometric}, and Spherical UNet~\cite{zhao2019spherical}, 
in brain phenotype prediction tasks have stressed these challenges.
% Print manuscripts two columns to a page, in the manner in which these
% instructions are printed. The exact dimensions for pages are:
% \begin{itemize}
%     \item left and right margins: .75$''$
%     \item column width: 3.375$''$
%     \item gap between columns: .25$''$
%     \item top margin---first page: 1.375$''$
%     \item top margin---other pages: .75$''$
%     \item bottom margin: 1.25$''$
%     \item column height---first page: 6.625$''$
%     \item column height---other pages: 9$''$
% \end{itemize}

% All measurements assume an 8-1/2$''$ $\times$ 11$''$ page size. For
% A4-size paper, use the given top and left margins, column width,
% height, and gap, and modify the bottom and right margins as necessary.

\subsection{Attention-based Methods}

The self-attention mechanism, introduced in the 
Transformer~\cite{vaswani2017attention}, has revolutionized natural 
language processing (NLP) by capturing long-range dependencies. 
This architecture become the foundation for models like 
BERT~\cite{devlin2018bert} and GPT~\cite{floridi2020gpt}. 
Researchers extended self-attention to visual representation 
learning by splitting image into patches, known as the Vision 
Transformer (ViT)~\cite{dosovitskiy2020image}. 
The Swin Transformer~\cite{liu2021swin} proposed hierarchical 
image merging via shifted windows, significantly improving 
efficiency and scalability for tasks like object detection 
and image segmentation. In medical image tasks, Surface 
Vision Transformer (SiT)~\cite{dahan2022surface} was proposed 
to address irregular geometries, such as the cerebral cortex. 
Motivated by the asymmetric development of brain structure, 
the hemispheric relation inference network (HRINet)~\cite{zhao2024attention} 
was designed to extract potential covariation relationships 
between bilateral hemispheres. However, the attention-based 
models not only struggle with quadratic time complexity relative 
to the sequence length, leading to significant computational 
costs in modeling dense, long sequences in NLP and high-resolution 
images in computer vision, but are also limited by quadratic 
space complexity owing to store all key-value pairs from 
previous sequences as its memory. To address these issues, 
numerous works have focused on reducing both quadratic time 
complexity and memory cost, as demonstrated 
in~\cite{wang2020linformer,qiu2019blockwise,han2023hyperattention,beltagy2020longformer,dao2022flashattention,dao2023flashattention,shah2024flashattention,han2025agent,han2024demystify} by 
changing the operation of attention calculation, but cause a drop in performance.

\subsection{State Space Models (SSMs)}

The SSMs have recently been proposed to address key limitations of 
Recurrent Neural Networks (RNNs)~\cite{medsker2001recurrent}, 
particularly the challenges of non-parallelizable training and the 
tendency to forget earlier information as sequence length increases. 
The structured state space for sequence (S4)~\cite{gu2021efficiently} 
employs the zero-order hold technique for discretization and 
High-order Polynomial Projection Operators (HiPPO)~\cite{gu2020hippo} to 
compress context into a smaller state. However, S4 is constrained by Linear 
Time Invariance (LTI), leading to limited ability to perform 
adaptive inference based on different inputs. In addition, S4 
fails to prioritize and attend to the most critical parts owing 
to treat each segment equally. Mamba~\cite{gu2023mamba} incorporates 
a selective scanning mechanism, allowing the model to 
selectively extract relevant information depending on the 
inputs. Vim~\cite{zhu2024vision} extended capabilities 
of Mamba to the computer vision domain, designing a generic vision 
backbone based on bidirectional SSM to address direction-sensitive 
problem, analogous to the ViT. Additionally, the Visual State Space 
Model (VMamba)~\cite{liu2024vmambavisualstatespace} proposed a 
cross-scan module to bridge the difference between 1D array scanning 
and 2D plane traversal, enabling the adaptation of Mamba for visual 
data while preserving the size of the receptive fields.

\begin{table*}[!ht]
    \centering
    % \begin{threeparttable}
    \begin{tabular}{l c c c c}
    \hline
        ~ & ~ & \textbf{\textit{dHCP}} & ~ & \textbf{\textit{Replication dataset}} \\ \cline{2-4}
        ~ & \textbf{\textit{Subset 1 (N=408)}} & \textbf{\textit{Subset 2 (N=16)}} & \textbf{\textit{Subset 3 (N=410)}} & \textbf{\textit{(N=10)}}  \\ \hline 
        \textbf{Birth age [weeks $^{\textbf{+days}}$]} & 39$^{+5}$ (39$^{+0}$ - 40$^{+6}$) & 30$^{+4}$ (28$^{+0}$ - 32$^{+4}$) & 38$^{+6}$ (38$^{+3}$ - 40$^{+5}$) & 37$^{+2}$ (35$^{+4}$ - 39$^{+0}$)  \\ 
        \textbf{Scan age [weeks $^{\textbf{+days}}$]} & 41$^{+0}$ (39$^{+5}$ - 42$^{+2}$) & 41$^{+4}$ (38$^{+3}$ - 43$^{+6}$) & 41$^{+1}$ (40$^{+0}$ - 42$^{+3}$) & 39$^{+4}$ (39$^{+2}$ - 40$^{+5}$) \\ 
        \textbf{Birth weight} & 3.28 (0.57) & 1.52 (0.62) & 3.11 (0.80) & 2.06 (1.00)  \\ 
        \textbf{Head circumference at scan} & 34.81 (1.84) & 34.85 (3.05) & 34.99 (1.81) & - \\ 
        \textbf{Radiology score (1/2/3/4/5)} & 254/154/0/0/0 & 3/3/5/1/4 & 203/129/47/9/22 & - \\ 
        \textbf{Gender (M/F)} & 225/183 & 9/7 & 210/200 & 6/11 \\ 
        \textbf{Language score} & - & - & 19.42 (5.26) & - \\ 
        \textbf{Motor score} & - & - & 20.55 (3.17) & - \\ \hline
    \end{tabular}
    % \begin{tablenotes}
    % \item[a] IQR: Interquartile Range.
    % \end{tablenotes}
    %  \end{threeparttable}
    \caption{Demographic and clinical information of the subjects.}
    \label{tab1}
\end{table*}

\section{Materials and Methods}
\subsection{Image acquisition and Dataset}
The imaging data used in this work are from the 
publicly available Developing Human Connectome 
Project (dHCP) and the Gansu Provincial Maternity 
and Child-care Hospital (GPMCH). We used T1-weighted (T1w) 
and T2-weighted (T2w) images to calculate morphometric 
metrics of cerebral cortex.

The dHCP is approved by the United Kingdom Health Research Ethics 
Authority (reference number: 14/LO/1169). Additionally, 
we collected T1w and T2w images of 10 infants from the 
GPMCH (2020-GSFY-05). These images were acquired in the 
resolution of 0.8$\times$0.8$\times$1.6 mm$^{3}$ with 0.8 mm 
overlap, and were reconstructed to 0.5 mm isotropic resolution. 

Concerning the data from dHCP, a total of 526 infants covering preterm- and term-born 
neonates ranging from 24 to 45 weeks postmenstrual age (PMA) are enrolled in our study. 
The neurodevelopmental assessments for these infants, conducted at 18 months of age 
using the Bayley-III Scales of Infant Development, can also be obtained. We used the 
following exclusion criteria: For PMA prediction, (\textit{i}) we excluded the later scans of 
participants who were scanned twice to avoid the influence of extrauterine environmental 
factors; (\textit{ii}) term-born neonates with focal abnormalities (radiology score \textgreater\ 2) were 
excluded to build a normative model for normal brain development assessment. The remaining 
infants were then split into two subsets: \textit{Subset 1}: 408 participants who were born and 
scanned between 34 and 45 PMA; \textit{Subset 2}: 16 preterm infants who were born before 34 
PMA and scanned at term-equivalent age (\textgreater\ 37 PMA) for the evaluation of 
premature effects on brain development. For scaled language and motor scores prediction, we 
retained the scans closest to 40 weeks for neonates scanned twice identified  
as \textit{Subset 3}: 410 infants born between 23 and 43 gestational weeks (GA). 
We further utilized data from GPMCH as a \textit{Replication dataset}, 
consisting of 10 neonates born and scanned 
between 34 and 40 PMA, to evaluate the generalization 
ability of the models. The demographic details are 
provided in Table \ref{tab1}\footnote{Birth age and scan age are using 
median (interquartile range) for presentation, while birth weight, head 
circumference at scan, language and motor score are using 
mean (standard deviation) for denoting. 14, 1 and 15 head circumference 
data were missed in \textit{Subset 1}, \textit{Subset 2}, 
and \textit{Subset 3}, respectively. The unit is centimeters.}.

Four cortical surface metrics—curvature, sulcal 
depth, cortical thickness and 
myelination (T1w/T2w ratio)—were used as 
features. Each feature channel was normalized 
using Z-score. \textit{Subset 1} and \textit{3} were 
split into training, validation, and testing 
datasets in an 8:1:1 ratio within each label interval. 
All data were registered to the dHCP 
40-week spherical template, which represents the 
cortical surface as an approximated sphere 
composed of triangles, with 32,492 vertices 
per hemisphere. We resampled the template sphere 
to a regular sixth-order icosphere (Ico-6) using 
barycentric interpolation.

\subsection{Preliminaries}
SSMs are generally considered LTI systems that map 
an input stimulation $u(t)\in\mathbb{R}^N$ to an output 
response $y(t)\in\mathbb{R}^N$ through a hidden state $h(t)\in\mathbb{R}^N$. 
The evolution of the hidden state over time is governed by parameter matrices
$\textbf{A}\in\mathbb{R}^{N \times N}$, 
$\textbf{B}\in\mathbb{R}^{N \times 1}$, 
and $\textbf{C}\in\mathbb{R}^{1 \times N}$.
The system is mathematically described using a linear ordinary 
differential equation (ODEs) as follows:
\begin{equation}\begin{cases}\label{eq1}
    & h^{\prime}(t)=\textbf{A}h(t) + \textbf{B}u(t) \\
    & y(t) = \textbf{C}h(t)
\end{cases}\end{equation}
where $\textbf{A}$ is the state matrix to control 
the latent state $h$, $\textbf{B}$ denotes the control matrix, 
and $\textbf{C}$ is the output matrix. 
Equation \eqref{eq1} aims to predict the state of the 
system based on observed data. Since the input is generally continuous, 
the primary use of SSMs is in continuous-time representation. However, 
since computers struggle with processing continuous signals and the real 
data we used is typically discrete, the standard 
procedure is to discretize Equation \eqref{eq1} using the Zero-order hold (ZOH) method  
which assumes that the input signal remains constant between sampling 
intervals, formulated as:
\begin{equation}\begin{cases}\label{1}
    & h_t=\overline{\textbf{A}}h_{t-1} + \textbf{B}u_t \\
    & y_t = \textbf{C}h_t
\end{cases}\end{equation}
where $\overline{\textbf{A}} = exp(\Delta \textbf{A})$ and $\overline{\textbf{B}} = (\Delta \textbf{A})^{-1}exp(\Delta \textbf{A} - I) \cdot \Delta \textbf{B}$
are the discretized parameter matrices and $\Delta$ is the discretization step size. The output $y$
is then calculated using a global convolution kernel $\overline{\textbf{K}}\in\mathbb{R}^L$, $L$ is the input sequence length. The kernel is defined as:
\begin{equation}\begin{cases}\label{1}
    & \overline{\textbf{K}}=(\textbf{C}\overline{\textbf{B}}, \textbf{C}\overline{\textbf{A}\textbf{B}}, \ldots, \textbf{C}\overline{\textbf{A}}^k\overline{\textbf{B}}, \ldots) \\
    & \textbf{y} = \textbf{u} \ast \overline{\textbf{K}}
\end{cases}\end{equation}
where $k\in[0, L)$ indicates the sequence index.

\begin{figure*}[!t]
    \centering
    \includegraphics[width=\textwidth]{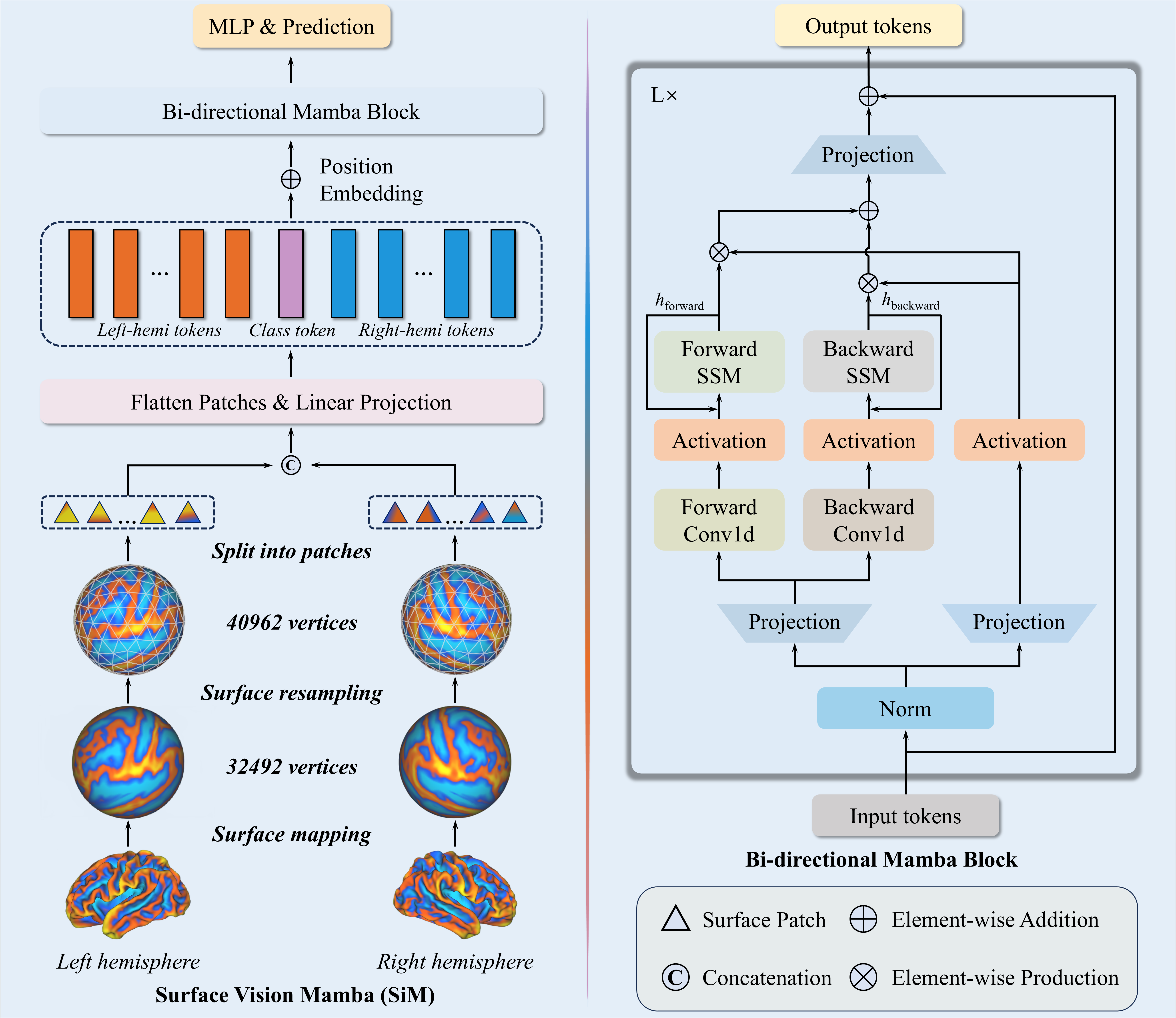}
    \caption{Overview of the proposed Surface Vision Mamba (SiM) architecture. The cortical data from the left and right hemispheres are initially mapped onto a 40-week spherical template with 32,492 vertices per hemisphere. The template is then resampled to a sixth-order icosphere containing 40,962 vertice, partitioned into triangular patches (taking Ico-2 shown as an example in the Figure \ref{fig1}) that fully cover the sphere. Surface patches from both hemispheres are concatenated, flattened, and linearly embedded. A learnable class token is inserted between the tokens from the left and right hemispheres, followed by the addition of positional embeddings. The processed data is then fed into the Bi-directional Mamba block.}
    \label{fig2}
\end{figure*}

\begin{table*}
    \centering
    \resizebox{0.9\linewidth}{!}{
        \begin{tabular}{lccccc}
            \toprule
            Icosphere Order  & First & Second & Third & Fourth & Fifth \\
            \midrule
            The number of patches (\textit{N}) & 80 & 320 & 1280 & 5120 & 20480 \\
            The number of vertices (\textit{V}) & 561 & 153 & 45 & 15 & 6 \\
            Input Dimension (\textit{VC}) & 2244 & 612 & 180 & 60 & 24 \\
            Sequence Length (\textit{2N}) & 160 & 640 & 2560 & 10240 & 40960 \\
            Icosphere Input Size & 160 $\times$ 2244 & 640 $\times$ 612 & 2560 $\times$ 180 & 10240 $\times$ 60 & 40960 $\times$ 24 \\
            \bottomrule
        \end{tabular}
    }
    \caption{Summary of the parameters for icospheres of different orders.}
    \label{tab2}
\end{table*}

\begin{table}
    \centering
    \setlength{\tabcolsep}{0.5mm}
    \fontsize{9}{11}\selectfont
    \resizebox{1\linewidth}{!}{
        \begin{tabular}{lcccc}
            \toprule
            Model  & Layers & Hidden size \textit{D} & Expanded size \textit{E} & Parameters \\
            \midrule
            SiM-Tiny & 24 & 192 & 384 & 7M \\
            SiM-Small & 24 & 384 & 768 & 24M \\
            SiM-Base & 24 & 768 & 1536 & 93M \\
            \bottomrule
        \end{tabular}
    }
    \caption{The configuration of different architecture variants.}
    \label{tab3}
\end{table}

\begin{table*}[ht]
\centering
\begin{tabular}{lcccccccccccc}
\toprule
\multirow{2}[2]{*}{\textbf{\textit{Methods}}} & 
\multicolumn{2}{c}{\textbf{\textit{Supervised}}} & 
\multicolumn{1}{c}{} & 
\multicolumn{2}{c}{\textbf{\textit{Fine-tuning}}} & 
\multicolumn{1}{c}{} & 
\multicolumn{2}{c}{\textbf{\textit{Autoregressive}}} & 
\multicolumn{1}{c}{\textbf{\textit{Params.}}} & 
\multicolumn{1}{c}{\textbf{\textit{MACs}}} \\
\cmidrule(lr){2-3} \cmidrule(lr){5-6} \cmidrule(lr){8-9}
& \textbf{MAE} & \textbf{MSE} & & \textbf{MAE} & \textbf{MSE} & & \textbf{MAE} & \textbf{MSE} & \textbf{(M)} & \textbf{(G)} \\ 
\toprule
\textbf{MoNet}              & 0.64±0.54 & 0.70±1.37 & & - & - & & - & - & - & -  \\
\textbf{S2CNN}              & 0.69±0.45 & 0.69±0.73 & & - & - & & - & - & - & -  \\
\textbf{ChebNet}            & 0.71±0.59 & 0.85±1.14 & & - & - & & - & - & - & -  \\
\textbf{GConvNet}           & 0.86±0.73 & 1.27±1.91 & & - & - & & - & - & - & -  \\
\textbf{PointNet++}         & 0.67±0.07 & 0.76±0.10 & & - & - & & - & - & - & -  \\
\textbf{Spherical UNet}     & 0.72±0.58 & 0.85±1.31 & & - & - & & - & - & - & -  \\
\toprule
\textbf{HRINet/1}           & 0.75±0.67 & 1.05±0.05 & & - & - & & - & - & 10 & - \\
\textbf{SiT-Tiny/1}         & 0.79±0.62 & 1.04±0.42 & & 0.81±0.59 & 1.37±0.46 & & - & - & 6 & 0.9  \\
\textbf{SiT-Small/1}        & 0.81±0.57 & 1.06±0.36 & & 0.87±0.63 & 0.93±0.28 & & - & - & 22 & 3.6 \\
\textbf{SiT-Base/1}         & 0.82±0.56 & 0.98±0.33 & & 0.86±0.74 & 1.13±0.47 & & - & - & 87 & 14.0 \\
\midrule
\textbf{SiM-Tiny/1}         & 0.85±0.64 & 0.91±0.31 & & 0.76±0.66 & 1.21±0.26 & & 1.03±0.70 & 1.87±0.42 & 7 & 1.9 \\
\textbf{SiM-Small/1}        & 0.87±0.74 & 1.60±0.38 & & 0.76±0.69 & 1.55±0.64 & & 1.26±0.83 & 2.43±0.20 & 24 & 4.2 \\
\textbf{SiM-Base/1}         & 0.86±0.69 & 1.20±0.04 & & 0.84±0.63 & 1.24±0.19 & & 1.06±0.67 & 1.34±0.30 & 92 & 15.4 \\
\toprule
\textbf{HRINet/2}           & 0.62±0.44 & \textbf{0.39±0.25} & & - & - & & - & - & 10 & - \\
\textbf{SiT-Tiny/2}         & 0.69±0.52 & 0.47±0.36 & & 0.66±0.58 & 0.78±0.01 & & - & - & 6 & 3.5 \\
\textbf{SiT-Small/2}        & 0.67±0.50 & 0.62±0.11 & & 0.65±0.46 & 0.65±0.03 & & - & - & 22 & 13.9 \\
\textbf{SiT-Base/2}         & 0.64±0.57 & 0.57±0.55 & & 0.72±0.47 & 0.58±0.20 & & - & - & 86 & 55.0 \\
\midrule
\textbf{SiM-Tiny/2}         & 1.09±0.80 & 2.00±0.22 & & 0.74±0.59 & 0.98±0.10 & & 1.04±0.77 & 2.19±0.65 & 6 & 4.7 \\
\textbf{SiM-Small/2}        & 0.98±0.80 & 1.89±0.38 & & 0.60±0.49 & 0.52±0.10 & & 1.12±0.92 & 3.50±1.79 & 24 & 16.5 \\
\textbf{SiM-Base/2}         & 0.88±0.66 & 1.27±0.09 & & 0.74±0.68 & 0.89±0.15 & & 1.20±0.83 & 3.17±1.33 & 91 & 61.5 \\
\toprule
\textbf{HRINet/3}           & \textit{OOM} & \textit{OOM} & & - & - & & - & - & 10 & - \\
\textbf{SiT-Tiny/3}         & 0.60±0.48 & 0.47±0.16 & & 0.62±0.50 & 0.53±0.13 & & - & - & 6 & 4.7 \\
\textbf{SiT-Small/3}        & 0.60±0.51 & 0.54±0.41 & & 0.60±0.43 & 0.42±0.16 & & - & - & 24 & 16.5 \\
\textbf{SiT-Base/3}         & \textit{OOM} & \textit{OOM} & & \textit{OOM} & \textit{OOM} & & - & - & 87 & 220.9 \\
\midrule
\textbf{SiM-Tiny/3}         & 1.09±0.76 & 2.79±1.30 & & 0.60±0.46 & 0.85±0.35 & & 0.91±0.81 & 3.62±2.73 & 7 & 18.9 \\
\textbf{SiM-Small/3}        & 1.09±0.84 & 1.78±0.14 & & \textbf{0.56±0.50} & 0.59±0.04 & & 0.87±0.65 & 1.89±0.91 & 24 & 66.1 \\
\textbf{SiM-Base/3}         & 1.03±0.81 & 2.30±0.73 & & 0.62±0.45 & 0.65±0.07 & & 1.09±0.71 & 2.27±0.76 & 93 & 245.5 \\
\bottomrule
\end{tabular}
\caption{Performance comparison on dHCP.}
\label{tab4}
\end{table*}

Mamba addressed key challenges in sequence modeling with a 
selective scanning mechanism and a faster hardware-aware algorithm. 
The former prioritizes and extracts the most significant information 
from sequence contexts, compressing it into a refined state 
and avoiding the inefficiency of treating all elements equally. 
The hardware-aware algorithm optimizes computational efficiency 
by utilizing modern hardware capabilities, enabling sub-quadratic 
time complexity. These characteristics make Mamba highly effective 
and scalable for processing long sequences while maintaining 
performance and computational efficiency.

\subsection{Surface Vision Mamba}
Given the interconnected nature of the brain, alterations within one region 
will inevitably influence others. To capture these long-range dependencies, 
we proposed the SiM model, as shown in Figure \ref{fig2}. Specifically, the input domain 
is divided into $2N$ patches, represented as $\widetilde{X} = \{\widetilde{L}, \widetilde{R}|\widetilde{L}\in\mathbb{R}^{N \times V \times C}, \widetilde{R}\in\mathbb{R}^{N \times V \times C}\}$, 
that $V$ is the number of vertices in a patch, and $C$ denotes the number of feature channels. This is then flattened to 
$X = \{L, R|L\in\mathbb{R}^{N\times(VC)}, R\in\mathbb{R}^{N\times(VC)}\}$. 
Next, we projected $X$ into \textit{D}-dimensional vectors using a trainable fully connected layer. Following the design of ViT and BERT, 
a learnable class token $X_{cls}$ is concatenated between the left and right hemispheres to represent the patch sequence. To retain positional information, 
standard 1D position embeddings $E_{pos}$ are added to the patch features.
\begin{equation}\label{1}
    S_0 = [X_L^1W; \cdots X_L^N;X_{cls};X_R^1W; \cdots;X_R^NW] + E_{pos} 
\end{equation}
where $W \in \mathbb{R}^{(VC)\times D}$, $E_{pos} \in \mathbb{R}^{(2N+1)\times D}$,
$S_0\in\mathbb{R}^{(2N+1)\times D}$ is the initial input of SiM, $X_L^1$ and $X_R^1$ 
represent the first patches of the left and right hemispheres, respectively. 
The implementation of SiM follows the same structure as Vim. Specifically, 
for a given layer $l$, the input from the previous layer
$S_{l-1}$ is processed as follows:
\begin{equation}\begin{cases}\label{1}
    & S_l = SiM(S_{l-1}) + S_{l-1} \\
    & T = LayerNorm(S_l^N) \\
    & \widehat{p} = MLP(T)
\end{cases}\end{equation}

\subsection{Surface Patching Methods}
The choice of surface patching methods can significantly affect model performance. 
In most \textit{Surface}-based visual tasks, each face of a second-order icosphere (Ico-2) 
is commonly used as a patch, with all data points in the face treated as vertices. 
This approach splits the surface into 320 non-overlapping patches, each 
containing 153 vertices, with patches sharing only common edges. As Mamba 
has been indicated to perform well on tasks involving long-sequences, we 
extend the sequence length by progressively subdividing the icosahedron into 
finer discrete levels and evaluate different surface patching methods, 
including first- to third-order icosphere, as summarized in Table \ref{tab2}. The 
icosphere subdivision process involves three steps: (\textit{i}) new vertices are 
inserted at the midpoints of edges from the previous subdivision level; 
(\textit{ii}) new edges are generated between adjacent new vertices within the same face; 
and (\textit{iii}) the newly added vertices are projected onto the circumsphere of the 
icosahedron. The different surface patching methods are visually represented 
in Figure \ref{fig1}. To evaluate the impact of patching methods, we use the Mean 
Absolute Error (MAE) as a performance metric and the Multiply-Accumulate 
Operations (MACs) is used to estimate computational cost.

\subsection{Training Methods} \label{3.5}
Medical imaging datasets are often smaller than natural imaging datasets 
due to ethical, privacy, and legal restrictions on image acquisition, the 
variability in imaging equipment and parameter settings, and the challenges 
of annotating large datasets. To overcome these limitations, pretraining 
methods are essential for learning robust features, enhancing performance 
on downstream tasks. In this study, we explore three training strategies: 
(\textit{i}) training models from scratch; (\textit{ii}) fine-tuning pretrained weights from 
ImageNet (as released in Vision Mamba); and (\textit{iii}) self-supervised pretraining 
for visual representation learning. Given the suitability of the Mamba 
architecture for autoregressive modeling~\cite{ren2024autoregressive}, we adopt autoregressive 
approach for self-supervised pretraining, where the model predicts the 
next token based on preceding information. The performance of all three 
training strategies is evaluated using mean squared error (MSE) as the 
loss function.

\section{Results and Discussion}
\subsection{Model Variants}
The proposed SiM configurations are built upon three variants of Vim: 
Vim-Tiny, Vim-Small, and Vim-Base. Table \ref{tab3} 
summarizes the architectural details of SiM. 
Concise notation is employed for model size and icosphere input size, e.g., 
SiM-B/3 represents the “Base” variant with an input size of 
2560$\times$180, using an Ico-3 grid on the sphere. 
With the constant resolution of Ico-6, higher-order icospheres yield 
finer-grained patches, enabling more precise analysis. This capability is 
peculiarly valuable in the medical imaging domain, where finer-grained 
insights are critical for capturing subtle variations related to diseases.

\subsection{Infant Brain Age Prediction}
As shown in Table \ref{tab4}\footnote{Case marked as (’-’) represents 
unspecified values. (’\textit{OOM}’) means out of memory. 
Bold intricates the best performance in MAE and MSE, 
respectively. The detailed setup is provided in A.1.}, the comparison 
results of SiM against benchmark GDL methods~\cite{fawaz2021benchmarking,vosylius2020geometric} and 
attention-based models~\cite{dahan2022surface,zhao2024attention} on PMA 
prediction of \textit{Subset 1} on the three training strategies (see section \ref{3.5}). 
Notably, when fine-tuning with ImageNet pretraining weights, three variants of SiM 
outperform all the GDL methods using an Ico-3 grid. With comparable model 
parameters and MACs, SiM-S/3 achieves a performance of 0.56±0.50, 
surpassing SiT-S/3 (0.60±0.43). However, when training from scratch, 
the performance of all SiM variants decreased obviously, likely 
reflecting a tendency to overfit on small datasets due to the absence of 
strong prior weights. Although self-supervised pretraining has been 
shown to effectively strengthen model performance in previous studies, 
this benefit is less evident in our results. Actually, only the SiM-T/3 
and SiM-S/3 exhibit improvements compared with training from scratch. 
This may be attributed to overfitting, which hampers generalization, 
or to the limited sample size, restricting the ability of model to 
capture sufficient features.

Additionally, we conducted further experiments:
\begin{enumerate}
\item Ablation studies on the decoder design in autoregressive pretraining, the best performance is achieved when the decoder depth is 1 and the width is 256, displayed in Table \ref{tab8}.

\item The impact of preterm birth on the brain development, we find that predicted brain age in preterm infants was significantly lower than chronological age in the \textit{Subset 2}, with \textit{MAE} = 0.89±0.87 and \textit{MSE} = 1.56±2.71, suggesting that preterm birth may delay brain development at term-equivalent age, shown in Figure \ref{fig5}.

\item Among all the models, SiM achieved the best performance in predicting Language (SiM-Base/2) and Motor (SiM-Small/1) outcomes at 18 months, with \textit{MAE} = 2.82±2.39, \textit{MSE} = 15.15±1.70, and \textit{MAE} = 1.55±1.17, \textit{MSE} = 2.32±1.76, respectively, both were achieved when training from scratch. The results are represented in Table \ref{tab9} and Table \ref{tab10}.

\item Generalization validation experiment. We validate the generalization performance of model using \textit{Replication dataset} in Table \ref{tab11}. The results show that our SiM-Small/3 exhibits the best generalization performance with \textit{MAE} = 1.17±0.95 and \textit{MSE} = 2.89±2.17 when training with ImageNet pretraining weights, despite all results experiencing some decline.
\end{enumerate}
The experiments mentioned above are provided in Appendix B and the parameter settings for all experiments are summarized in Appendix A.

\begin{figure*}[!t]
    \centering
    \includegraphics[width=\textwidth]{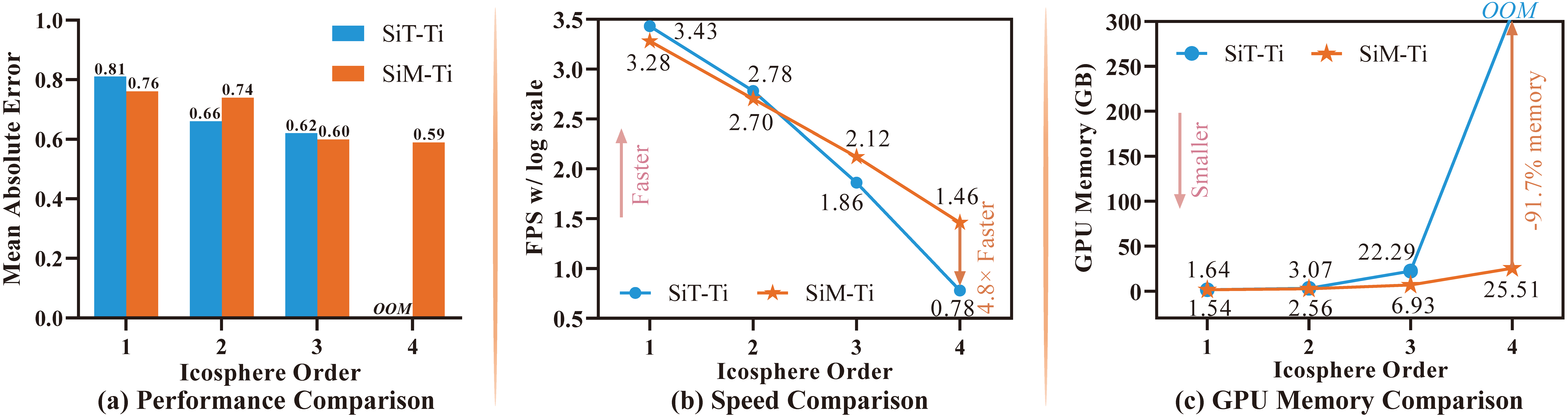}
    \caption{Comparison of PMA prediction performance and efficiency between SiT and our SiM.}
    \label{fig3}
\end{figure*}

\begin{figure}[!t]
    \centering
    \includegraphics[width=\columnwidth]{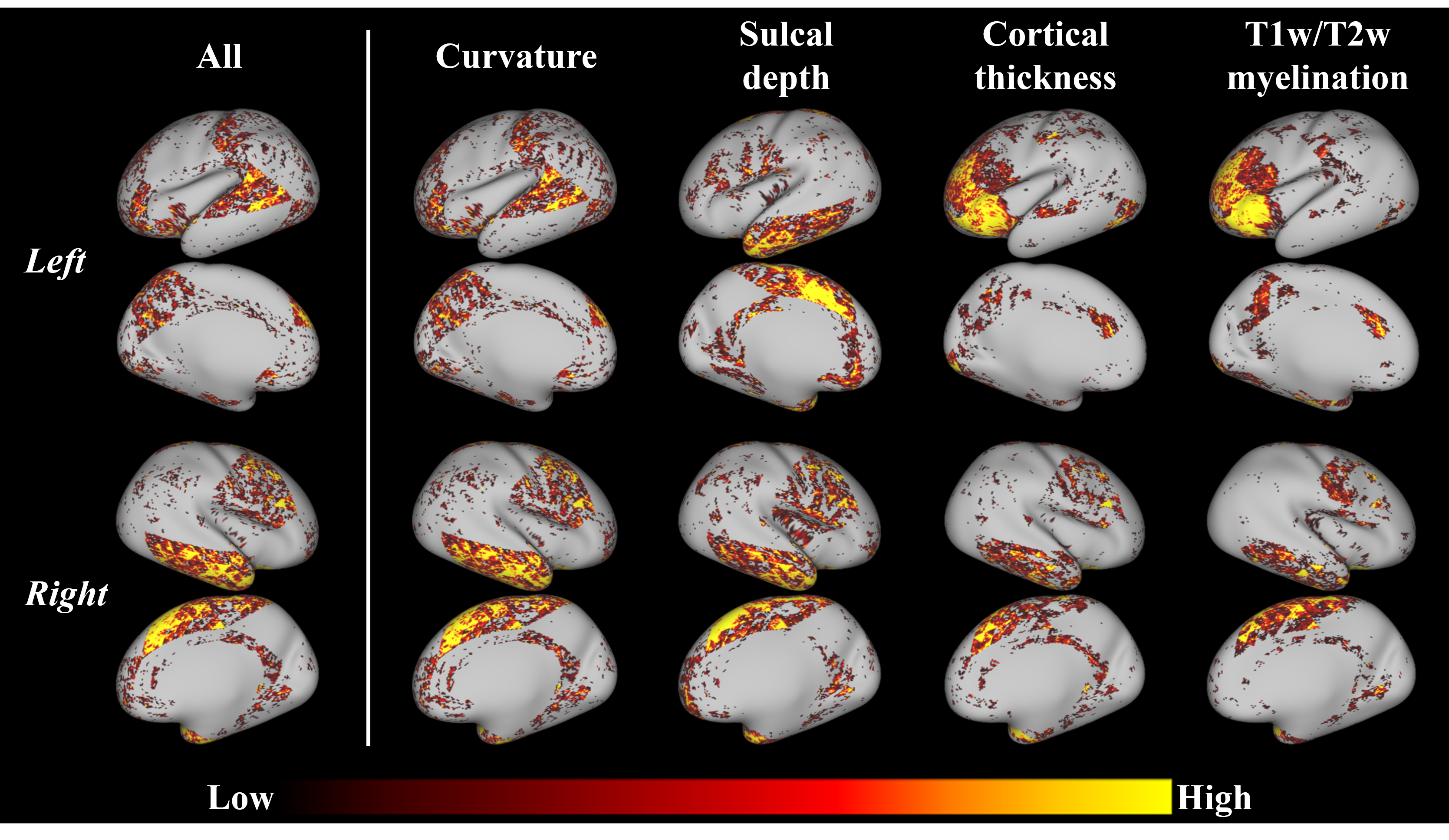}
    \caption{Spatial distribution of informative vertices for PMA prediction.}
    \label{fig4}
\end{figure}
\subsection{Long Sequence and Efficiency Analysis}
Figure \ref{fig3} illustrates the performance and efficiency of the tiny-sized 
SiM model across different surface patching methods. In terms of MAE, SiM 
slightly outperforms SiT across all other patching methods except when using 
an Ico-2 grid, and both models exhibit decreasing MAE as the icosphere order 
increases (Figure \ref{fig3}a). Regarding FPS (Frames Per Second), SiM is 
slightly slower than SiT when the icophere order is below 3 but surpasses 
it as the order increases (Figure \ref{fig3}b). For GPU memory usage, 
SiM exhibits better efficiency with SiT when icosphere orders raises. 
Notably, when using an Ico-4 as grid, SiM is 4.8 times faster and 
consumes 91.7\% less GPU memory compared to SiT (Figure \ref{fig3}c). 
All experiments on efficiency analysis are conducted on a 40G A100 device. 
These results highlight the suitability of SiM for finer-grained tasks 
and its potential for practical clinical applications.

\subsection{Cortical Regions with Significant Contributions to Age Prediction}
We perform a sensitive analysis~\cite{saltelli2002sensitivity} on the test dataset 
of \textit{Subset 1} to evaluate the contribution of individual vertex on 
cortical surface to brain age prediction, as illustrated in Figure \ref{fig4}. 
For each vertex on the brain surface, we assessed four morphometric 
features (i.e., curvature, sulcal depth, cortical thickness, and myelination). 
“All” nullifies all features, while other results nullify one feature at a 
time per vertex. Contributions were quantified by measuring performance changes 
before and after nullification, with larger shifts indicating greater impact.

The mean performance change for each vertex/feature was computed, 
normalized using Z-score, and visualized on the cortical surface (Figure \ref{fig4}). 
The intensity of the hot color signifies the influences of each vertex/feature. 
In the right hemisphere, key regions include the temporal lobe, precentral 
gyrus, and prefrontal and paracentral cortices. The left hemisphere shows a 
similar focus on the prefrontal cortex, sensory cortex, language areas, and 
parietal cortex when all features or only curvature are masked. Sulcal depth 
complements these findings by emphasizing the temporal lobe, central sulcus 
regions, and superior frontal and parietal areas. Cortical thickness and 
myelination highlight frontal regions, particularly the anterior insula, 
rostral middle frontal gyrus, and orbitofrontal cortex. 

\section{Conclusion}
In this study, we introduced Surface Vision Mamba (SiM), a novel vision 
backbone with sub-quadratic time complexity, tailored for genus-zero surfaces. 
We validated SiM as a more robust and efficient alternative to SiT in the 
challenging task of neurodevelopmental phenotype prediction from cortical 
surface data. Leveraging the strengths of Mamba in handling \textit{long-sequence} 
and \textit{autoregressive modeling}, we extended sequence lengths using various 
surface patching methods and conducted autoregressive pretraining. While 
SiM demonstrated sensitivity to sequence length, the benefits of autoregressive 
pretraining were limited, likely due to constraints of small samples. 
The use of longer sequences facilitated finer-grained partitioning, 
enhancing the ability to identify potential pathological features critical 
in clinical applications. Furthermore, SiM offers faster inference speeds 
and lower GPU memory consumption, making it both efficient and practical. 
Sensitivity analysis also emphasized the interpretability of SiM, 
highlighting its potential utility in medical research and applications.

\appendix

\section*{Ethical Statement}

There are no ethical issues.

% \newpage
\bibliographystyle{named}
\bibliography{ijcai25}

\begin{thebibliography}{}

\bibitem[\protect\citeauthoryear{Beltagy \bgroup \em et al.\egroup }{2020}]{beltagy2020longformer}
Iz~Beltagy, Matthew~E Peters, and Arman Cohan.
\newblock Longformer: The long-document transformer.
\newblock {\em arXiv preprint arXiv:2004.05150}, 2020.

\bibitem[\protect\citeauthoryear{Dahan \bgroup \em et al.\egroup }{2022}]{dahan2022surface}
Simon Dahan, Abdulah Fawaz, Logan~ZJ Williams, Chunhui Yang, Timothy~S Coalson, Matthew~F Glasser, A~David Edwards, Daniel Rueckert, and Emma~C Robinson.
\newblock Surface vision transformers: Attention-based modelling applied to cortical analysis.
\newblock In {\em International Conference on Medical Imaging with Deep Learning}, pages 282--303. PMLR, 2022.

\bibitem[\protect\citeauthoryear{Dao \bgroup \em et al.\egroup }{2022}]{dao2022flashattention}
Tri Dao, Dan Fu, Stefano Ermon, Atri Rudra, and Christopher R{\'e}.
\newblock Flashattention: Fast and memory-efficient exact attention with io-awareness.
\newblock {\em Advances in Neural Information Processing Systems}, 35:16344--16359, 2022.

\bibitem[\protect\citeauthoryear{Dao}{2023}]{dao2023flashattention}
Tri Dao.
\newblock Flashattention-2: Faster attention with better parallelism and work partitioning.
\newblock {\em arXiv preprint arXiv:2307.08691}, 2023.

\bibitem[\protect\citeauthoryear{Devlin}{2018}]{devlin2018bert}
Jacob Devlin.
\newblock Bert: Pre-training of deep bidirectional transformers for language understanding.
\newblock {\em arXiv preprint arXiv:1810.04805}, 2018.

\bibitem[\protect\citeauthoryear{Dosovitskiy}{2020}]{dosovitskiy2020image}
Alexey Dosovitskiy.
\newblock An image is worth 16x16 words: Transformers for image recognition at scale.
\newblock {\em arXiv preprint arXiv:2010.11929}, 2020.

\bibitem[\protect\citeauthoryear{Fawaz \bgroup \em et al.\egroup }{2021}]{fawaz2021benchmarking}
Abdulah Fawaz, Logan~ZJ Williams, Amir Alansary, Cher Bass, Karthik Gopinath, Mariana da~Silva, Simon Dahan, Chris Adamson, Bonnie Alexander, Deanne Thompson, et~al.
\newblock Benchmarking geometric deep learning for cortical segmentation and neurodevelopmental phenotype prediction.
\newblock {\em bioRxiv}, pages 2021--12, 2021.

\bibitem[\protect\citeauthoryear{Floridi and Chiriatti}{2020}]{floridi2020gpt}
Luciano Floridi and Massimo Chiriatti.
\newblock Gpt-3: Its nature, scope, limits, and consequences.
\newblock {\em Minds and Machines}, 30:681--694, 2020.

\bibitem[\protect\citeauthoryear{Gu and Dao}{2023}]{gu2023mamba}
Albert Gu and Tri Dao.
\newblock Mamba: Linear-time sequence modeling with selective state spaces.
\newblock {\em arXiv preprint arXiv:2312.00752}, 2023.

\bibitem[\protect\citeauthoryear{Gu \bgroup \em et al.\egroup }{2020}]{gu2020hippo}
Albert Gu, Tri Dao, Stefano Ermon, Atri Rudra, and Christopher R{\'e}.
\newblock Hippo: Recurrent memory with optimal polynomial projections.
\newblock {\em Advances in neural information processing systems}, 33:1474--1487, 2020.

\bibitem[\protect\citeauthoryear{Gu \bgroup \em et al.\egroup }{2021}]{gu2021efficiently}
Albert Gu, Karan Goel, and Christopher R{\'e}.
\newblock Efficiently modeling long sequences with structured state spaces.
\newblock {\em arXiv preprint arXiv:2111.00396}, 2021.

\bibitem[\protect\citeauthoryear{Han \bgroup \em et al.\egroup }{2022}]{han2022vision}
Kai Han, Yunhe Wang, Jianyuan Guo, Yehui Tang, and Enhua Wu.
\newblock Vision gnn: An image is worth graph of nodes.
\newblock {\em Advances in neural information processing systems}, 35:8291--8303, 2022.

\bibitem[\protect\citeauthoryear{Han \bgroup \em et al.\egroup }{2023}]{han2023hyperattention}
Insu Han, Rajesh Jayaram, Amin Karbasi, Vahab Mirrokni, David~P Woodruff, and Amir Zandieh.
\newblock Hyperattention: Long-context attention in near-linear time.
\newblock {\em arXiv preprint arXiv:2310.05869}, 2023.

\bibitem[\protect\citeauthoryear{Han \bgroup \em et al.\egroup }{2024}]{han2024demystify}
Dongchen Han, Ziyi Wang, Zhuofan Xia, Yizeng Han, Yifan Pu, Chunjiang Ge, Jun Song, Shiji Song, Bo~Zheng, and Gao Huang.
\newblock Demystify mamba in vision: A linear attention perspective.
\newblock {\em arXiv preprint arXiv:2405.16605}, 2024.

\bibitem[\protect\citeauthoryear{Han \bgroup \em et al.\egroup }{2025}]{han2025agent}
Dongchen Han, Tianzhu Ye, Yizeng Han, Zhuofan Xia, Siyuan Pan, Pengfei Wan, Shiji Song, and Gao Huang.
\newblock Agent attention: On the integration of softmax and linear attention.
\newblock In {\em European Conference on Computer Vision}, pages 124--140. Springer, 2025.

\bibitem[\protect\citeauthoryear{Liu \bgroup \em et al.\egroup }{2021}]{liu2021swin}
Ze~Liu, Yutong Lin, Yue Cao, Han Hu, Yixuan Wei, Zheng Zhang, Stephen Lin, and Baining Guo.
\newblock Swin transformer: Hierarchical vision transformer using shifted windows.
\newblock In {\em Proceedings of the IEEE/CVF international conference on computer vision}, pages 10012--10022, 2021.

\bibitem[\protect\citeauthoryear{Liu \bgroup \em et al.\egroup }{2024}]{liu2024vmambavisualstatespace}
Yue Liu, Yunjie Tian, Yuzhong Zhao, Hongtian Yu, Lingxi Xie, Yaowei Wang, Qixiang Ye, Jianbin Jiao, and Yunfan Liu.
\newblock Vmamba: Visual state space model, 2024.

\bibitem[\protect\citeauthoryear{Medsker \bgroup \em et al.\egroup }{2001}]{medsker2001recurrent}
Larry~R Medsker, Lakhmi Jain, et~al.
\newblock Recurrent neural networks.
\newblock {\em Design and Applications}, 5(64-67):2, 2001.

\bibitem[\protect\citeauthoryear{Monti \bgroup \em et al.\egroup }{2017}]{monti2017geometric}
Federico Monti, Davide Boscaini, Jonathan Masci, Emanuele Rodola, Jan Svoboda, and Michael~M Bronstein.
\newblock Geometric deep learning on graphs and manifolds using mixture model cnns.
\newblock In {\em Proceedings of the IEEE conference on computer vision and pattern recognition}, pages 5115--5124, 2017.

\bibitem[\protect\citeauthoryear{Qiu \bgroup \em et al.\egroup }{2019}]{qiu2019blockwise}
Jiezhong Qiu, Hao Ma, Omer Levy, Scott Wen-tau Yih, Sinong Wang, and Jie Tang.
\newblock Blockwise self-attention for long document understanding.
\newblock {\em arXiv preprint arXiv:1911.02972}, 2019.

\bibitem[\protect\citeauthoryear{Ren \bgroup \em et al.\egroup }{2024}]{ren2024autoregressive}
Sucheng Ren, Xianhang Li, Haoqin Tu, Feng Wang, Fangxun Shu, Lei Zhang, Jieru Mei, Linjie Yang, Peng Wang, Heng Wang, et~al.
\newblock Autoregressive pretraining with mamba in vision.
\newblock {\em arXiv preprint arXiv:2406.07537}, 2024.

\bibitem[\protect\citeauthoryear{Saltelli}{2002}]{saltelli2002sensitivity}
Andrea Saltelli.
\newblock Sensitivity analysis for importance assessment.
\newblock {\em Risk analysis}, 22(3):579--590, 2002.

\bibitem[\protect\citeauthoryear{Shah \bgroup \em et al.\egroup }{2024}]{shah2024flashattention}
Jay Shah, Ganesh Bikshandi, Ying Zhang, Vijay Thakkar, Pradeep Ramani, and Tri Dao.
\newblock Flashattention-3: Fast and accurate attention with asynchrony and low-precision.
\newblock {\em arXiv preprint arXiv:2407.08608}, 2024.

\bibitem[\protect\citeauthoryear{Vaswani}{2017}]{vaswani2017attention}
A~Vaswani.
\newblock Attention is all you need.
\newblock {\em Advances in Neural Information Processing Systems}, 2017.

\bibitem[\protect\citeauthoryear{Vosylius \bgroup \em et al.\egroup }{2020}]{vosylius2020geometric}
Vitalis Vosylius, Andy Wang, Cemlyn Waters, Alexey Zakharov, Francis Ward, Loic Le~Folgoc, John Cupitt, Antonios Makropoulos, Andreas Schuh, Daniel Rueckert, et~al.
\newblock Geometric deep learning for post-menstrual age prediction based on the neonatal white matter cortical surface.
\newblock In {\em Uncertainty for Safe Utilization of Machine Learning in Medical Imaging, and Graphs in Biomedical Image Analysis: Second International Workshop, UNSURE 2020, and Third International Workshop, GRAIL 2020, Held in Conjunction with MICCAI 2020, Lima, Peru, October 8, 2020, Proceedings 2}, pages 174--186. Springer, 2020.

\bibitem[\protect\citeauthoryear{Wang \bgroup \em et al.\egroup }{2020}]{wang2020linformer}
Sinong Wang, Belinda~Z Li, Madian Khabsa, Han Fang, and Hao Ma.
\newblock Linformer: Self-attention with linear complexity.
\newblock {\em arXiv preprint arXiv:2006.04768}, 2020.

\bibitem[\protect\citeauthoryear{Yu and Wang}{2024}]{yu2024mambaout}
Weihao Yu and Xinchao Wang.
\newblock Mambaout: Do we really need mamba for vision?
\newblock {\em arXiv preprint arXiv:2405.07992}, 2024.

\bibitem[\protect\citeauthoryear{Zhao \bgroup \em et al.\egroup }{2019}]{zhao2019spherical}
Fenqiang Zhao, Shunren Xia, Zhengwang Wu, Dingna Duan, Li~Wang, Weili Lin, John~H Gilmore, Dinggang Shen, and Gang Li.
\newblock Spherical u-net on cortical surfaces: methods and applications.
\newblock In {\em Information Processing in Medical Imaging: 26th International Conference, IPMI 2019, Hong Kong, China, June 2--7, 2019, Proceedings 26}, pages 855--866. Springer, 2019.

\bibitem[\protect\citeauthoryear{Zhao \bgroup \em et al.\egroup }{2024}]{zhao2024attention}
Leilei Zhao, Dalin Zhu, Xiaomin Wang, Xia Liu, Tongtong Li, Boyang Wang, Zhijun Yao, Weihao Zheng, and Bin Hu.
\newblock An attention-based hemispheric relation inference network for perinatal brain age prediction.
\newblock {\em IEEE Journal of Biomedical and Health Informatics}, 2024.

\bibitem[\protect\citeauthoryear{Zhu \bgroup \em et al.\egroup }{2024}]{zhu2024vision}
Lianghui Zhu, Bencheng Liao, Qian Zhang, Xinlong Wang, Wenyu Liu, and Xinggang Wang.
\newblock Vision mamba: Efficient visual representation learning with bidirectional state space model.
\newblock {\em arXiv preprint arXiv:2401.09417}, 2024.

\end{thebibliography}

\clearpage

\begin{table*}[ht]
\centering
\setlength{\tabcolsep}{2.5mm}
\begin{tabular}{l*{3}{>{\centering\arraybackslash}p{0.7cm}}*{3}{>{\centering\arraybackslash}p{0.7cm}}*{3}{>{\centering\arraybackslash}p{0.7cm}}*{3}{>{\centering\arraybackslash}p{0.9cm}}}
\toprule
\multirow{2}{*}{} & 
\multicolumn{3}{c}{\textit{Scratch}} & 
\multicolumn{3}{c}{\textit{Fine-tuning}} & 
\multicolumn{3}{c}{\textit{AR Pretraining}} & 
\multicolumn{3}{c}{\textit{AR Fine-tuning}} \\
\cmidrule(lr){2-4} \cmidrule(lr){5-7} \cmidrule(lr){8-10} \cmidrule(lr){11-13}
& T & S & B & T & S & B & T & S & B & T & S & B \\
\midrule
Epochs          & \multicolumn{3}{c}{1000} & \multicolumn{3}{c}{600} & 4000 & 3000 & 3000 & \multicolumn{3}{c}{600} \\
Batch size      & \multicolumn{3}{c}{32} & \multicolumn{3}{c}{32} & \multicolumn{3}{c}{32} & \multicolumn{3}{c}{32} \\
Optimizer       & \multicolumn{3}{c}{AdamW} & \multicolumn{3}{c}{AdamW} & \multicolumn{3}{c}{AdamW} & \multicolumn{3}{c}{AdamW} \\
Adam $\epsilon$ & \multicolumn{3}{c}{1e-8} & \multicolumn{3}{c}{1e-8} & \multicolumn{3}{c}{1e-8} & \multicolumn{3}{c}{1e-8} \\
Adam ($\beta_1$, $\beta_2$) & \multicolumn{3}{c}{(0.9, 0.999)} & \multicolumn{3}{c}{(0.9, 0.999)} & \multicolumn{3}{c}{(0.9, 0.999)} & \multicolumn{3}{c}{(0.9, 0.999)} \\
LR              & \multicolumn{3}{c}{5e-5} & 5e-5 & 5e-5 & 3e-5 & \multicolumn{3}{c}{1.5e-4} & 1.5e-4 & 1e-4 & 8e-5 \\
LR decay        & \multicolumn{3}{c}{Linear} & \multicolumn{3}{c}{Linear} & \multicolumn{3}{c}{Cosine} & \multicolumn{3}{c}{Cosine} \\
Step size       & \multicolumn{3}{c}{500} & \multicolumn{3}{c}{200} & \multicolumn{3}{c}{-} & \multicolumn{3}{c}{-} \\
Gamma           & \multicolumn{3}{c}{0.5} & \multicolumn{3}{c}{0.5} & \multicolumn{3}{c}{-} & \multicolumn{3}{c}{-} \\
Gradient clipping & \multicolumn{3}{c}{None} & \multicolumn{3}{c}{None} & \multicolumn{3}{c}{None} & \multicolumn{3}{c}{None} \\
Warmup epochs   & \multicolumn{3}{c}{None} & \multicolumn{3}{c}{None} & \multicolumn{3}{c}{10} & \multicolumn{3}{c}{10} \\
Weight decay    & \multicolumn{3}{c}{1e-8} & \multicolumn{3}{c}{1e-8} & \multicolumn{3}{c}{0.5} & \multicolumn{3}{c}{1e-6} \\
EMA decay rate  & \multicolumn{3}{c}{None} & \multicolumn{3}{c}{None} & \multicolumn{3}{c}{None} & \multicolumn{3}{c}{None} \\
\bottomrule
\end{tabular}
\caption{Hyperparameters for all training strategies in PMA prediction. Specifically, scratch means that training from scratch, fine-tuning refers to using ImageNet pretraining weights in Vim. For self-supervised pre-training, AR means autoregressive. T, S, B represent tiny-size, small-size, base-size, respectively.}
\label{tab5}
\end{table*}

\begin{table*}[ht]
\centering
\setlength{\tabcolsep}{2.5mm}
\begin{tabular}{l*{3}{>{\centering\arraybackslash}p{0.7cm}}*{3}{>{\centering\arraybackslash}p{0.7cm}}*{3}{>{\centering\arraybackslash}p{0.7cm}}*{3}{>{\centering\arraybackslash}p{0.9cm}}}
\toprule
\multirow{2}{*}{} & 
\multicolumn{3}{c}{\textit{Scratch}} & 
\multicolumn{3}{c}{\textit{Fine-tuning}} & 
\multicolumn{3}{c}{\textit{AR Pretraining}} & 
\multicolumn{3}{c}{\textit{AR Fine-tuning}} \\
\cmidrule(lr){2-4} \cmidrule(lr){5-7} \cmidrule(lr){8-10} \cmidrule(lr){11-13}
& T & S & B & T & S & B & T & S & B & T & S & B \\
\midrule
Epochs          & 200 & 200 & 100 & \multicolumn{3}{c}{100} & 4000 & 3000 & 3000 & \multicolumn{3}{c}{300} \\
Batch size      & \multicolumn{3}{c}{32} & \multicolumn{3}{c}{32} & \multicolumn{3}{c}{32} & \multicolumn{3}{c}{32} \\
Optimizer       & \multicolumn{3}{c}{AdamW} & \multicolumn{3}{c}{AdamW} & \multicolumn{3}{c}{AdamW} & \multicolumn{3}{c}{AdamW} \\
Adam $\epsilon$ & \multicolumn{3}{c}{1e-8} & \multicolumn{3}{c}{1e-8} & \multicolumn{3}{c}{1e-8} & \multicolumn{3}{c}{1e-8} \\
Adam ($\beta_1$, $\beta_2$) & \multicolumn{3}{c}{(0.9, 0.999)} & \multicolumn{3}{c}{(0.9, 0.999)} & \multicolumn{3}{c}{(0.9, 0.999)} & \multicolumn{3}{c}{(0.9, 0.999)} \\
LR              & 8e-5 & 6e-5 & 6e-5 & 8e-5 & 8e-5 & 6e-5 & \multicolumn{3}{c}{1.5e-4} & 1.8e-4 & 1.2e-4 & 8e-5 \\
LR decay        & \multicolumn{3}{c}{Cosine} & \multicolumn{3}{c}{Cosine} & \multicolumn{3}{c}{Cosine} & \multicolumn{3}{c}{Cosine} \\
Step size       & \multicolumn{3}{c}{-} & \multicolumn{3}{c}{-} & \multicolumn{3}{c}{-} & \multicolumn{3}{c}{-} \\
Gamma           & \multicolumn{3}{c}{-} & \multicolumn{3}{c}{-} & \multicolumn{3}{c}{-} & \multicolumn{3}{c}{-} \\
Gradient clipping & \multicolumn{3}{c}{None} & \multicolumn{3}{c}{None} & \multicolumn{3}{c}{None} & \multicolumn{3}{c}{None} \\
Warmup epochs   & \multicolumn{3}{c}{10} & \multicolumn{3}{c}{10} & \multicolumn{3}{c}{10} & \multicolumn{3}{c}{10} \\
Weight decay    & \multicolumn{3}{c}{1e-4} & \multicolumn{3}{c}{1e-4} & \multicolumn{3}{c}{0.5} & \multicolumn{3}{c}{1e-6} \\
EMA decay rate  & \multicolumn{3}{c}{None} & \multicolumn{3}{c}{None} & \multicolumn{3}{c}{None} & \multicolumn{3}{c}{None} \\
\bottomrule
\end{tabular}
\caption{Hyperparameters for all training strategies in Scaled Language Score prediction.}
\label{tab6}
\end{table*}

\begin{table*}[ht]
\centering
\setlength{\tabcolsep}{2.5mm}
\begin{tabular}{l*{3}{>{\centering\arraybackslash}p{0.7cm}}*{3}{>{\centering\arraybackslash}p{0.7cm}}*{3}{>{\centering\arraybackslash}p{0.7cm}}*{3}{>{\centering\arraybackslash}p{0.9cm}}}
\toprule
\multirow{2}{*}{} & 
\multicolumn{3}{c}{\textit{Scratch}} & 
\multicolumn{3}{c}{\textit{Fine-tuning}} & 
\multicolumn{3}{c}{\textit{AR Pretraining}} & 
\multicolumn{3}{c}{\textit{AR Fine-tuning}} \\
\cmidrule(lr){2-4} \cmidrule(lr){5-7} \cmidrule(lr){8-10} \cmidrule(lr){11-13}
& T & S & B & T & S & B & T & S & B & T & S & B \\
\midrule
Epochs          & \multicolumn{3}{c}{300} & \multicolumn{3}{c}{200} & 4000 & 3000 & 3000 & \multicolumn{3}{c}{200} \\
Batch size      & \multicolumn{3}{c}{32} & \multicolumn{3}{c}{32} & \multicolumn{3}{c}{32} & \multicolumn{3}{c}{32} \\
Optimizer       & \multicolumn{3}{c}{AdamW} & \multicolumn{3}{c}{AdamW} & \multicolumn{3}{c}{AdamW} & \multicolumn{3}{c}{AdamW} \\
Adam $\epsilon$ & \multicolumn{3}{c}{1e-8} & \multicolumn{3}{c}{1e-8} & \multicolumn{3}{c}{1e-8} & \multicolumn{3}{c}{1e-8} \\
Adam ($\beta_1$, $\beta_2$) & \multicolumn{3}{c}{(0.9, 0.999)} & \multicolumn{3}{c}{(0.9, 0.999)} & \multicolumn{3}{c}{(0.9, 0.999)} & \multicolumn{3}{c}{(0.9, 0.999)} \\
LR              & 1e-4 & 8e-5 & 8e-5 & 6e-5 & 4e-5 & 4e-5 & \multicolumn{3}{c}{1.5e-4} & 2e-4 & 1.5e-4 & 1.2e-4 \\
LR decay        & \multicolumn{3}{c}{Cosine} & \multicolumn{3}{c}{Cosine} & \multicolumn{3}{c}{Cosine} & \multicolumn{3}{c}{Cosine} \\
Step size       & \multicolumn{3}{c}{-} & \multicolumn{3}{c}{-} & \multicolumn{3}{c}{-} & \multicolumn{3}{c}{-} \\
Gamma           & \multicolumn{3}{c}{-} & \multicolumn{3}{c}{-} & \multicolumn{3}{c}{-} & \multicolumn{3}{c}{-} \\
Gradient clipping & \multicolumn{3}{c}{None} & \multicolumn{3}{c}{None} & \multicolumn{3}{c}{None} & \multicolumn{3}{c}{None} \\
Warmup epochs   & 50 & 10 & 50 & \multicolumn{3}{c}{10} & \multicolumn{3}{c}{10} & 50 & 10 & 10 \\
Weight decay    & \multicolumn{3}{c}{1e-4} & \multicolumn{3}{c}{1e-4} & \multicolumn{3}{c}{0.5} & \multicolumn{3}{c}{1e-6} \\
EMA decay rate  & \multicolumn{3}{c}{None} & \multicolumn{3}{c}{None} & \multicolumn{3}{c}{None} & \multicolumn{3}{c}{None} \\
\bottomrule
\end{tabular}
\caption{Hyperparameters for all training strategies in Scaled Motor Score prediction.}
\label{tab7}
\end{table*}

\twocolumn

\section*{Appendix A. Implementation Details} \label{Appendix A}

All the experiments are implemented with 
Python 3.10.13 and PyTorch library and 
conducted on 4 NVIDIA A100 GPUs with batch 
size of 32. The Vim-Tiny\textsuperscript{†}, Vim-Small\textsuperscript{†} and 
Vim-Base weights are adapted to initialize 
our model that Vim-Tiny\textsuperscript{†} and Vim-Small\textsuperscript{†} are 
fine-tuned under long sequence but no 
open-source Vim-Base\textsuperscript{†} to use. The autoregressive 
pretraining for all model variants is performed 
using an Ico-3 grid and fine-tuned on the 
sequence obtained from other partitioning methods.

\subsection*{A.1. PMA Prediction Experiments}
We assess the performance of our models in PMA 
prediction using the \textit{Subset 1} which is described 
in Table \ref{tab1}. For all the model variants and surface 
patching methods, all training details were 
presented explicitly in Table \ref{tab5}.

\subsection*{A.2. Scaled Language Score Prediction Experiments}
Predicting long-term language scores is a 
challenging task, directly applying the PMA 
recipe does not work. We find that it is no need 
for a long schedule and provide our configuration in Table \ref{tab6}.

\subsection*{A.3. Scaled Motor Score Prediction Experiments}
The training setting is in Table \ref{tab7} and we 
observe that this to be akin to language 
score prediction. It is worth noting that a 
large warmup epoch can improve performance in some cases.

\section*{Appendix B. Additional Results} \label{Appendix B}
\subsection*{B.1. Decoder Design}
We use SiM-T/3 for decoder architectures 
ablation experiments on \textit{subset 1}, the fine-tuning 
MAE and MSE are summarized in Table \ref{tab8}. We first 
vary the \textit{decoder depth}, and find that deeper 
decoders decreased performance. 
Further increase the \textit{decoder width}, 
we find that performance improved when the dimension 
increased to 256, and then decreased.

\begin{table}
    \centering
    \begin{tabular}{cccc}
        \toprule
        blocks  & dim & MAE       & MSE \\
        \midrule
        1       & 256 & \textbf{0.91±0.81} & \textbf{3.62±2.73} \\
        2       & 256 & 0.97±0.81 & 3.54±2.51 \\
        4       & 256 & 1.08±0.97 & 2.99±1.15 \\
        \midrule
        1       & 128 & 1.04±0.87 & 3.34±1.94 \\
        1       & 256 & \textbf{0.91±0.81} & \textbf{3.62±2.73} \\
        1       & 512 & 1.05±0.83 & 3.88±2.68 \\
        \bottomrule
    \end{tabular}
    \caption{Ablation on decoder architectures. The reconstruction target is normalized features of each vertex. Default settings are marked in bold.}
    \label{tab8}
\end{table}

\begin{figure}[!t]
    \centering
    \includegraphics[width=\columnwidth]{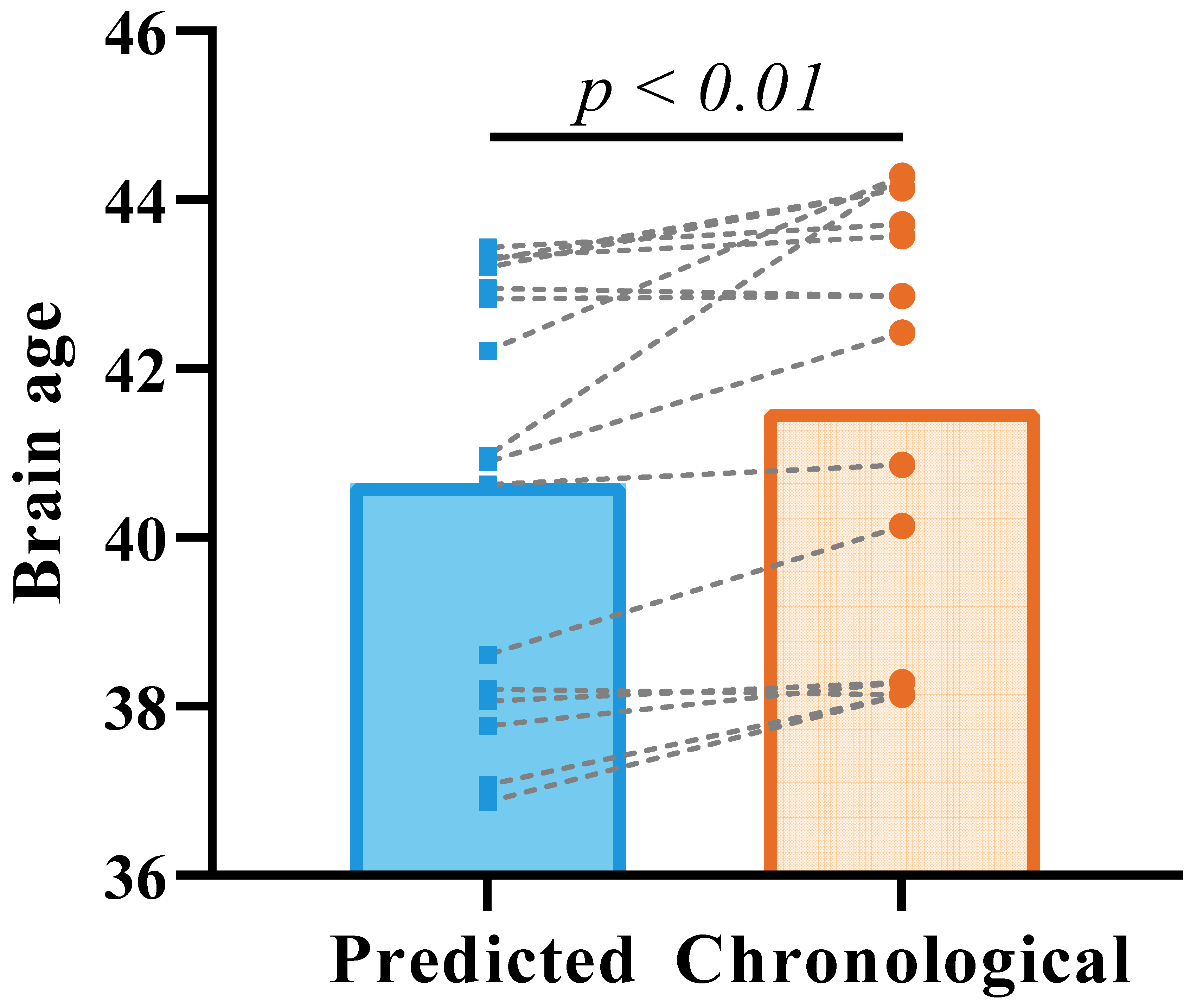}
    \caption{The significant difference between predicted and chronological brain age in preterm infants.}
    \label{fig5}
\end{figure}

\subsection*{B.2. Brain Development Analysis of Preterm Infant}
To explore whether preterm birth affects brain 
development, we applied the 
SiM-S/3 (fine-tuned with ImageNet pretraining weights) to 
the \textit{Subset 2}. The paired samples t-test was utilized 
to assess the difference between predicted brain 
age and chronological brain age. The result showed 
that predicted brain age in preterm infants was 
significantly lower than chronological age in the 
\textit{Subset 2}, with \textit{MAE} = 0.89±0.87 and \textit{MSE} = 1.56±2.71, 
as shown in Figure \ref{fig5} (\textit{p} $<$ 0.01). The finding suggest 
that preterm birth may delay brain development at 
term-equivalent age.

\subsection*{B.3. Infant Scaled Language and Motor Score Prediction}
Table \ref{tab9} and Table \ref{tab10} shows the prediction results 
for language and motor scores, respectively. The 
proposed SiM achieved the best performance compared 
to other methods.

\subsection*{B.4. Generalization Validation}
We validated the generalization for different 
models on the \textit{Replication dataset}. The results 
indicate the performance of all the models 
decreased, but our SiM still demonstrated the 
best generalization performance with \textit{MAE} = 1.17±0.95 
and \textit{MSE} = 2.89±2.17. All results are detailed 
in Table \ref{tab11}.

\begin{table*}[ht]
\centering
\begin{tabular}{lccccccccccc}
\toprule
\multirow{2}[2]{*}{\textbf{\textit{Methods}}} & 
\multicolumn{2}{c}{\textbf{\textit{Supervised}}} & 
\multicolumn{1}{c}{} & 
\multicolumn{2}{c}{\textbf{\textit{Fine-tuning}}} & 
\multicolumn{1}{c}{} & 
\multicolumn{2}{c}{\textbf{\textit{Autoregressive}}} \\
\cmidrule(lr){2-3} \cmidrule(lr){5-6} \cmidrule(lr){8-9}
& \textbf{MAE} & \textbf{MSE} & & \textbf{MAE} & \textbf{MSE} & & \textbf{MAE} & \textbf{MSE} \\ 
\toprule
\textbf{MoNet}              & 3.13±2.04 & \textbf{13.94±15.14} & & - & - & & - & -  \\
\textbf{S2CNN}              & 3.16±2.04 & 14.15±15.86 & & - & - & & - & -  \\
\textbf{ChebNet}            & 3.56±2.21 & 17.59±18.50 & & - & - & & - & -  \\
\textbf{GConvNet}           & 3.18±2.07 & 14.40±16.68 & & - & - & & - & -  \\
\textbf{PointNet++}         & 3.45±0.32 & 18.54±3.55  & & - & - & & - & -  \\
\textbf{Spherical UNet}     & 3.00±2.68 & 16.17±27.48 & & - & - & & - & -  \\
\toprule
\textbf{HRINet/1}           & 3.38±2.09 & 29.48±15.50 & & - & - & & - & - \\
\textbf{SiT-Tiny/1}         & 3.12±2.06 & 22.70±9.90  & & 3.27±2.07 & 28.05±14.85 & & - & -  \\
\textbf{SiT-Small/1}        & 3.25±2.04 & 24.67±11.24 & & 3.31±1.96 & 21.86±8.00 & & - & - \\
\textbf{SiT-Base/1}         & 3.23±2.07 & 23.86±10.35 & & 3.32±2.22 & 23.71±8.80 & & - & - \\
\midrule
\textbf{SiM-Tiny/1}         & 3.22±2.29 & 21.34±6.48  & & 3.12±2.30 & 15.74±0.82 & & 3.55±2.38 & 29.62±12.82 \\
\textbf{SiM-Small/1}        & 2.99±2.21 & 16.20±2.67  & & 2.99±2.06 & 16.99±3.55 & & 3.25±2.11 & 14.42±0.70 \\
\textbf{SiM-Base/1}         & 2.91±2.47 & 13.63±1.02  & & 3.05±2.15 & 17.83±4.44 & & 3.28±2.19 & 18.84±3.77 \\
\toprule
\textbf{HRINet/2}           & 3.38±2.10 & 29.95±15.97 & & - & - & & - & - \\
\textbf{SiT-Tiny/2}         & 3.22±2.06 & 28.50±12.09 & & 3.41±2.01 & 28.64±14.70 & & - & - \\
\textbf{SiT-Small/2}        & 3.35±2.20 & 32.50±20.28 & & 3.25±1.99 & 24.55±11.38 & & - & - \\
\textbf{SiT-Base/2}         & 3.11±2.03 & 22.84±10.30 & & 3.38±2.07 & 28.52±14.56 & & - & - \\
\midrule
\textbf{SiM-Tiny/2}         & 3.23±2.30 & 21.44±6.54  & & 3.02±2.13 & 16.94±3.74 & & 3.40±2.12 & 29.75±15.58 \\
\textbf{SiM-Small/2}        & 3.19±2.25 & 15.52±4.87  & & 3.02±2.23 & 18.66±5.18 & & 3.39±2.17 & 32.27±18.23 \\
\textbf{SiM-Base/2}         & \textbf{2.82±2.39} & 15.15±1.70 & & 3.12±2.15 & 18.22±4.38 & & 3.41±2.39 & 30.33±14.74 \\
\toprule
\textbf{HRINet/3}           & \textit{OOM} & \textit{OOM} & & - & - & & - & - \\
\textbf{SiT-Tiny/3}         & 3.20±2.06 & 23.44±10.15 & & 3.24±2.04 & 24.23±10.82 & & - & - \\
\textbf{SiT-Small/3}        & 3.20±2.06 & 23.20±9.85 & & 3.24±2.13 & 25.70±12.06 & & - & - \\
\textbf{SiT-Base/3}         & \textit{OOM} & \textit{OOM} & & \textit{OOM} & \textit{OOM} & & - & - \\
\midrule
\textbf{SiM-Tiny/3}         & 3.04±2.14 & 17.64±4.29 & & 3.03±2.13 & 17.35±4.07 & & 3.36±2.12 & 29.74±15.81 \\
\textbf{SiM-Small/3}        & 3.05±2.15 & 17.76±4.38 & & 2.98±2.23 & 15.60±1.99 & & 3.49±1.99 & 27.17±12.52 \\
\textbf{SiM-Base/3}         & 3.02±2.00 & 14.91±2.02 & & 2.90±2.30 & 16.49±3.19 & & 3.46±2.43 & 19.37±1.67 \\
\bottomrule
\end{tabular}
\caption{Infant Scaled Language Score Prediction on \textit{subset 3}.}
\label{tab9}
\end{table*}

\begin{table*}[ht]
\centering
\begin{tabular}{lccccccccccc}
\toprule
\multirow{2}[2]{*}{\textbf{\textit{Methods}}} & 
\multicolumn{2}{c}{\textbf{\textit{Supervised}}} & 
\multicolumn{1}{c}{} & 
\multicolumn{2}{c}{\textbf{\textit{Fine-tuning}}} & 
\multicolumn{1}{c}{} & 
\multicolumn{2}{c}{\textbf{\textit{Autoregressive}}} \\
\cmidrule(lr){2-3} \cmidrule(lr){5-6} \cmidrule(lr){8-9}
& \textbf{MAE} & \textbf{MSE} & & \textbf{MAE} & \textbf{MSE} & & \textbf{MAE} & \textbf{MSE} \\ 
\toprule
\textbf{MoNet}              & 2.07±1.52 & 6.60±8.89 & & - & - & & - & -  \\
\textbf{S2CNN}              & 1.75±1.20 & 4.49±5.64 & & - & - & & - & -  \\
\textbf{ChebNet}            & 1.99±1.52 & 6.26±8.35 & & - & - & & - & -  \\
\textbf{GConvNet}           & 2.10±1.62 & 7.04±9.52 & & - & - & & - & -  \\
\textbf{PointNet++}         & 2.19±0.02 & 6.23±0.86 & & - & - & & - & -  \\
\textbf{Spherical UNet}     & 2.04±1.53 & 6.49±8.40 & & - & - & & - & -  \\
\toprule
\textbf{HRINet/1}           & 1.78±1.27 & 2.72±2.50 & & - & - & & - & - \\
\textbf{SiT-Tiny/1}         & 1.83±1.28 & 3.04±2.35 & & 1.73±1.30 & 3.41±1.52 & & - & -  \\
\textbf{SiT-Small/1}        & 1.78±1.26 & 2.77±2.39 & & 1.88±1.29 & 3.27±2.31 & & - & - \\
\textbf{SiT-Base/1}         & 1.80±1.25 & 2.93±2.25 & & 1.76±1.26 & 2.84±2.20 & & - & - \\
\midrule
\textbf{SiM-Tiny/1}         & 1.75±1.19 & 2.57±2.33 & & 1.63±1.19 & 2.35±2.07 & & 1.74±1.25 & 2.62±2.40 \\
\textbf{SiM-Small/1}        & \textbf{1.55±1.17} & \textbf{2.32±1.76} & & 1.66±1.28 & 2.53±2.28 & & 1.83±1.23 & 3.78±1.32 \\
\textbf{SiM-Base/1}         & 1.56±1.22 & 2.52±1.71 & & 1.80±1.32 & 2.82±2.58 & & 1.72±1.31 & 3.10±1.90 \\
\toprule
\textbf{HRINet/2}           & 1.77±1.27 & 2.71±2.49 & & - & - & & - & - \\
\textbf{SiT-Tiny/2}         & 1.73±1.26 & 2.71±2.25 & & 1.88±1.27 & 3.14±2.42 & & - & - \\
\textbf{SiT-Small/2}        & 1.80±1.28 & 2.97±2.31 & & 1.78±1.25 & 2.86±2.26 & & - & - \\
\textbf{SiT-Base/2}         & 1.81±1.27 & 3.00±2.30 & & 1.97±1.21 & 3.79±1.86 & & - & - \\
\midrule
\textbf{SiM-Tiny/2}         & 1.99±1.47 & 3.58±3.05 & & 1.80±1.26 & 2.75±2.51 & & 1.84±1.39 & 3.07±2.68 \\
\textbf{SiM-Small/2}        & 1.84±1.33 & 3.19±2.38 & & 1.78±1.42 & 3.57±1.95 & & 1.94±1.50 & 3.38±3.18 \\
\textbf{SiM-Base/2}         & 1.83±1.37 & 2.98±2.70 & & 1.78±1.31 & 2.78±2.55 & & 1.93±1.44 & 3.98±2.18 \\
\toprule
\textbf{HRINet/3}           & \textit{OOM} & \textit{OOM} & & - & - & & - & - \\
\textbf{SiT-Tiny/3}         & 1.78±1.25 & 2.87±2.25 & & 1.82±1.27 & 2.86±2.49 & & - & - \\
\textbf{SiT-Small/3}        & 1.80±1.25 & 2.96±2.25 & & 1.76±1.26 & 2.88±2.18 & & - & - \\
\textbf{SiT-Base/3}         & \textit{OOM} & \textit{OOM} & & \textit{OOM} & \textit{OOM} & & - & - \\
\midrule
\textbf{SiM-Tiny/3}         & 2.00±1.68 & 3.75±3.73 & & 1.79±1.27 & 2.94±2.24 & & 1.80±1.33 & 2.84±2.63 \\
\textbf{SiM-Small/3}        & 1.82±1.39 & 2.90±2.83 & & 1.81±1.30 & 2.79±2.61 & & 1.78±1.29 & 2.74±2.51 \\
\textbf{SiM-Base/3}         & 1.84±1.40 & 3.01±2.82 & & 1.70±1.25 & 2.54±2.31 & & 1.92±1.45 & 3.27±3.04 \\
\bottomrule
\end{tabular}
\caption{Infant Scaled Motor Score Prediction on \textit{subset 3}.}
\label{tab10}
\end{table*}

\begin{table*}[ht]
\centering
\begin{tabular}{lccccccccccc}
\toprule
\multirow{2}[2]{*}{\textbf{\textit{Methods}}} & 
\multicolumn{2}{c}{\textbf{\textit{Supervised}}} & 
\multicolumn{1}{c}{} & 
\multicolumn{2}{c}{\textbf{\textit{Fine-tuning}}} & 
\multicolumn{1}{c}{} & 
\multicolumn{2}{c}{\textbf{\textit{Autoregressive}}} \\
\cmidrule(lr){2-3} \cmidrule(lr){5-6} \cmidrule(lr){8-9}
& \textbf{MAE} & \textbf{MSE} & & \textbf{MAE} & \textbf{MSE} & & \textbf{MAE} & \textbf{MSE} \\ 
\toprule
\textbf{MoNet}              & 1.64±0.89 & 3.09±2.43 & & - & - & & - & -  \\
\textbf{S2CNN}              & 1.83±1.21 & 4.65±6.09 & & - & - & & - & -  \\
\textbf{ChebNet}            & 1.58±1.40 & 4.45±6.52 & & - & - & & - & -  \\
\textbf{GConvNet}           & 1.98±1.07 & 5.08±3.75 & & - & - & & - & -  \\
\textbf{PointNet++}         & 3.27±1.07 & 13.04±6.39 & & - & - & & - & -  \\
\textbf{Spherical UNet}     & 1.86±1.04 & 2.78±3.12 & & - & - & & - & -  \\
\toprule
\textbf{HRINet/1}           & 8.61±1.88 & 77.63±30.35 & & - & - & & - & - \\
\textbf{SiT-Tiny/1}         & 3.70±1.54 & 16.08±10.24 & & 8.43±2.11 & 75.55±32.83 & & - & -  \\
\textbf{SiT-Small/1}        & 3.41±1.34 & 13.40±9.51 & & 4.02±1.09 & 17.33±8.98 & & - & - \\
\textbf{SiT-Base/1}         & 4.00±1.68 & 18.84±12.48 & & 6.19±1.65 & 41.05±17.92 & & - & - \\
\midrule
\textbf{SiM-Tiny/1}         & 4.32±1.75 & 20.05±9.68 & & 5.72±2.16 & 36.81±11.64 & & 3.85±1.62 & 15.74±7.50 \\
\textbf{SiM-Small/1}        & 5.32±1.66 & 28.42±11.30 & & 3.60±1.62 & 14.19±6.33 & & 2.05±1.41 & 7.42±4.46 \\
\textbf{SiM-Base/1}         & 4.81±1.69 & 24.03±10.03 & & 3.78±2.97 & 20.06±21.66 & & 1.63±1.47 & 6.59±6.33 \\
\toprule
\textbf{HRINet/2}           & 4.25±1.74 & 21.08±12.31 & & - & - & & - & - \\
\textbf{SiT-Tiny/2}         & 2.90±1.32 & 10.13±7.32 & & 4.88±1.84 & 27.16±14.99 & & - & - \\
\textbf{SiT-Small/2}        & 3.33±1.38 & 13.01±8.89 & & 2.08±0.78 & 4.95±3.31 & & - & - \\
\textbf{SiT-Base/2}         & 2.72±1.45 & 9.50±8.43 & & 4.73±1.75 & 25.49±13.24 & & - & - \\
\midrule
\textbf{SiM-Tiny/2}         & 4.48±2.62 & 28.31±12.52 & & 5.04±1.83 & 26.66±8.56 & & 6.11±2.24 & 39.59±14.42 \\
\textbf{SiM-Small/2}        & 4.66±1.77 & 23.88±8.53 & & 2.01±0.96 & 5.52±2.14 & & 4.12±1.63 & 18.06±8.10 \\
\textbf{SiM-Base/2}         & 4.81±1.69 & 24.03±10.03 & & 1.51±1.50 & 6.08±5.48 & & 3.43±1.55 & 12.92±5.04 \\
\toprule
\textbf{HRINet/3}           & - & - & & - & - & & - & - \\
\textbf{SiT-Tiny/3}         & 2.83±1.49 & 10.26±8.43 & & 2.18±1.02 & 5.79±4.28 & & - & - \\
\textbf{SiT-Small/3}        & 2.91±1.44 & 10.53±8.37 & & 1.86±0.99 & 4.43±3.89 & & - & - \\
\textbf{SiT-Base/3}         & - & - & & - & - & & - & - \\
\midrule
\textbf{SiM-Tiny/3}         & 5.44±1.72 & 25.09±12.40 & & 1.56±0.97 & 4.11±2.61 & & 3.54±1.24 & 13.09±4.92 \\
\textbf{SiM-Small/3}        & 4.66±1.77 & 23.88±8.53 & & \textbf{1.17±0.95} & \textbf{2.89±2.17} & & 3.30±1.67 & 13.25±5.05 \\
\textbf{SiM-Base/3}         & 4.36±1.76 & 21.48±7.48 & & 1.61±1.08 & 3.65±0.89 & & 4.26±1.89 & 19.40±8.06 \\
\bottomrule
\end{tabular}
\caption{Generalization Validation on \textit{Replication dataset}.}
\label{tab11}
\end{table*}

\end{document}